\pdfoutput=1

\documentclass[10pt,twocolumn,letterpaper]{article}

\usepackage{cvpr}              

\usepackage{xcolor}

\definecolor{iccvblue}{rgb}{0.21,0.49,0.74}

\usepackage{colortbl}
\usepackage{float}
\usepackage{graphicx}
\usepackage{multirow}
\usepackage{adjustbox}
\usepackage{arydshln}

\usepackage[pagebackref,breaklinks,colorlinks,allcolors=iccvblue]{hyperref}

\definecolor{upcolor}{RGB}{57,182,74}

\definecolor{cvprblue}{rgb}{0.21,0.49,0.74}

\def\paperID{} 
\def\confName{CVPR}
\def\confYear{2026}

\title{PaCo-FR: Patch-Pixel Aligned End-to-End Codebook Learning\\ for Facial Representation Pre-training}

\author{%
Yin Xie$^1$ \quad Zhichao Chen$^1$ \quad Zeyu Xiao$^2$ \quad Yongle Zhao$^1$ \quad Xiang An$^1$ \quad Kaicheng Yang$^1$ \\
Zimin Ran$^3$ \quad Jia Guo$^4$ \quad Ziyong Feng$^1$ \quad Jiankang Deng$^5$\thanks{Corresponding author} \\
$^1$ GlintLab \quad $^2$ National University of Singapore \quad \\ $^3$ University of Technology Sydney \quad $^4$ InsightFace.AI  \quad $^5$ Imperial College London \\
{\tt\small yiye.xieyin@gmail.com, jiankangdeng@gmail.com}
}

\begin{document}
\maketitle
\begin{abstract}
Facial representation pre-training is crucial for tasks like facial recognition, expression analysis, and virtual reality. 
However, existing methods face three key challenges: (1) failing to capture distinct facial features and fine-grained semantics, (2) ignoring the spatial structure inherent to facial anatomy, and (3) inefficiently utilizing limited labeled data. 
To overcome these, we introduce PaCo-FR, an unsupervised framework that combines masked image modeling with patch-pixel alignment.
Our approach integrates three innovative components: 
(1) a structured masking strategy that preserves spatial coherence by aligning with semantically meaningful facial regions, 
(2) a novel patch-based codebook that enhances feature discrimination with multiple candidate tokens, 
and (3) spatial consistency constraints that preserve geometric relationships between facial components.
PaCo-FR achieves state-of-the-art performance across several facial analysis tasks with just 2 million unlabeled images for pre-training. 
Our method demonstrates significant improvements, particularly in scenarios with varying poses, occlusions, and lighting conditions. 
We believe this work advances facial representation learning and offers a scalable, efficient solution that reduces reliance on expensive annotated datasets, driving more effective facial analysis systems.
\end{abstract}
\vspace{-4mm}    
\section{Introduction}
\label{sec:intro}

\begin{figure}[!t]
  \centering
  \includegraphics[width=\linewidth]{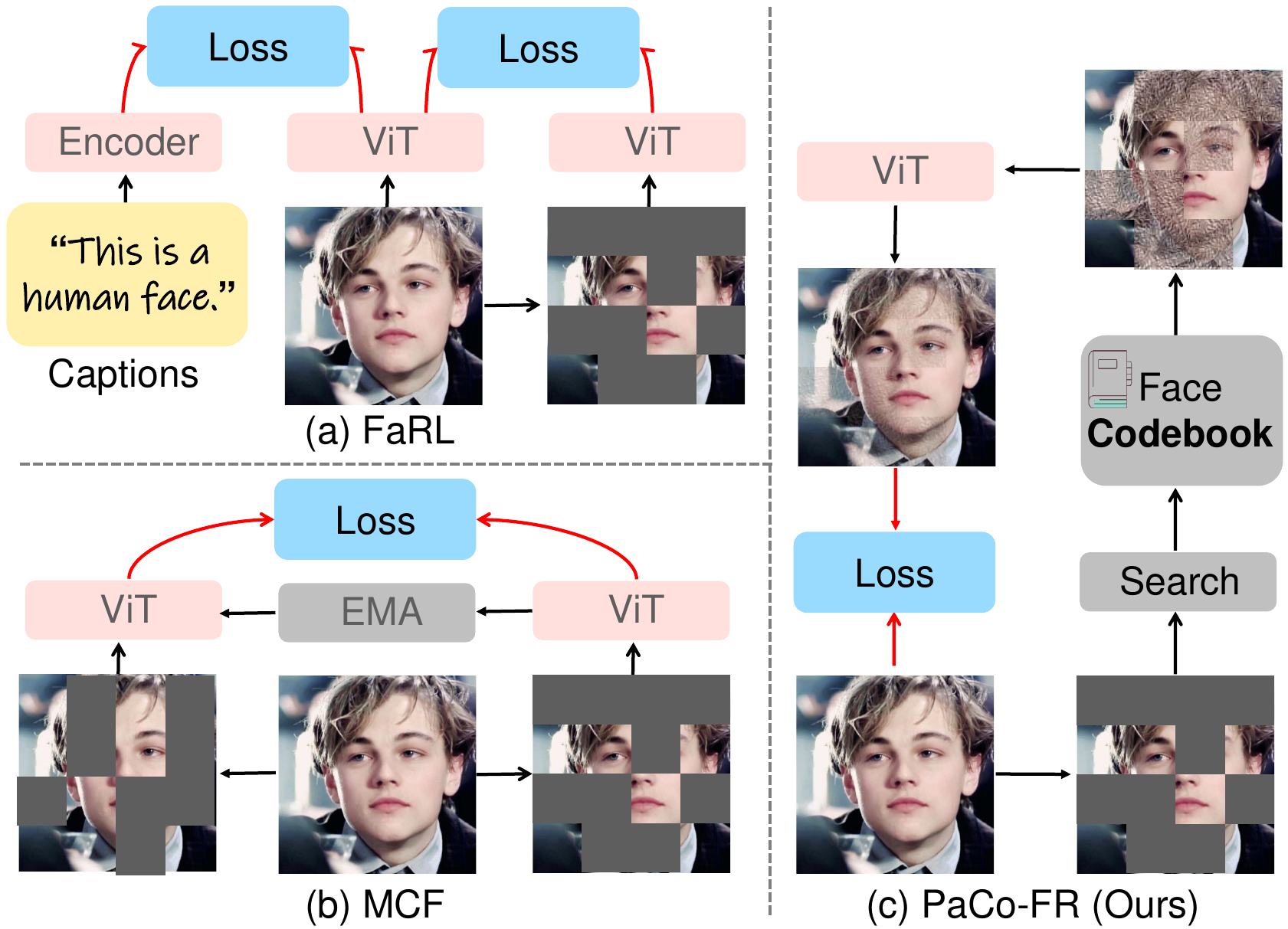}
  \vspace{-6mm}
  \caption{(a) The pipeline of FaRL~\cite{zheng2022general}, (b) The pipeline of MCF~\cite{wang2023toward}, and (c) our proposed PaCo-FR.
  PaCo-FR is specifically designed for facial characteristics and optimize it as a learning objective for fatial representation pre-training.
 }
  \label{tees}
  \vspace{-5mm}
\end{figure}

Facial representation learning is fundamental to human-centric AI~\cite{bulat2022pre,zheng2022general,shi2020towards,gao2024self}, supporting applications such as face recognition~\cite{zhao2003face}, identity verification~\cite{robertson2019face}, expression analysis~\cite{becattini2022understanding}, and avatar animation~\cite{zhang2023magicavatar}. The quality of facial representations affects the robustness and fairness of downstream systems. With facial imagery proliferating on social media, there is an urgent need for representations that are accurate, generalizable, and efficient for deployment.

Meanwhile, large-scale visual pre-training has become the standard in computer vision~\cite{abnar2021exploring}, aided by methods like MoCo~\cite{he2020momentum,chen2020improved,fan2021multiscale}, SimCLR~\cite{chen2020simple}, and CLIP~\cite{radford2021learning}. These approaches excel in extracting general visual features using unsupervised or weakly supervised learning on massive datasets. However, when applied to facial tasks, such models often fall short because they lack inductive alignment with the unique structures and semantics of human faces, pointing to the need for pre-training strategies tailored to facial data.

Recent research has started to address this gap with domain-specific approaches. FaRL~\cite{zheng2022general} introduced a large face-centric dataset (LAION-FACE) and vision-language pre-training, while MCF~\cite{wang2023toward} applied contrastive learning on a curated subset, showing the benefits of face-specific strategies. Nonetheless, fully exploiting the spatial regularities and fine-grained semantic details of faces remains a challenge.

\begin{figure}[!t]
  \centering
  \begin{subfigure}[b]{0.48\textwidth}
    \centering
    \includegraphics[width=\linewidth]{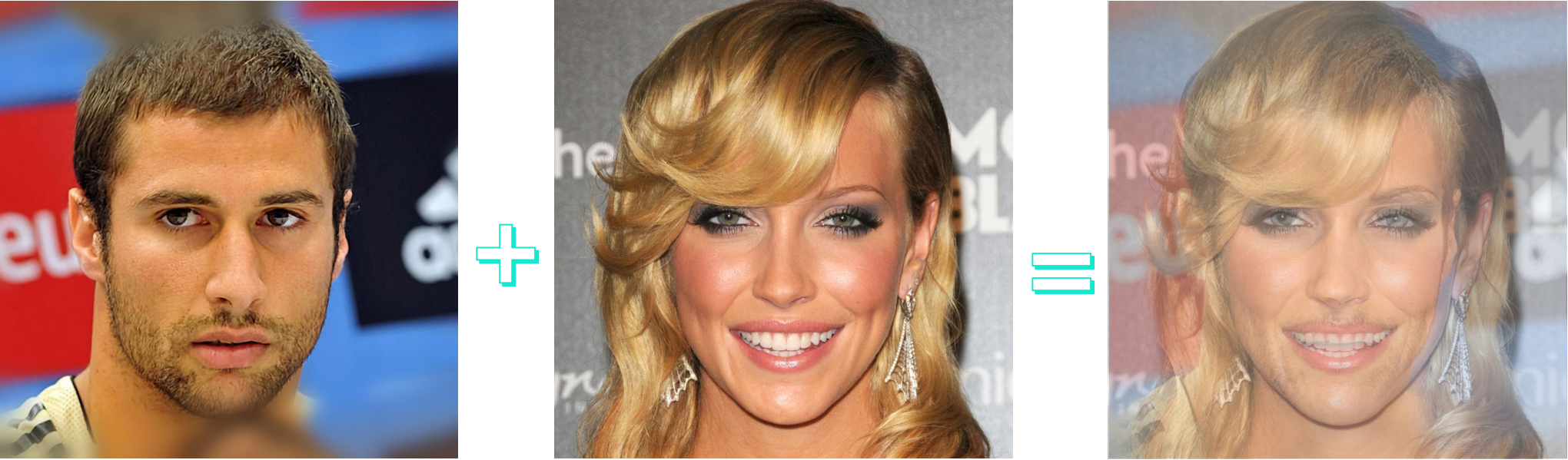}
    \caption{Face alignment.}
    \label{fig:face_align}
  \end{subfigure}
  \hfill
  \begin{subfigure}[b]{0.48\textwidth}
    \centering
    \includegraphics[width=\linewidth]{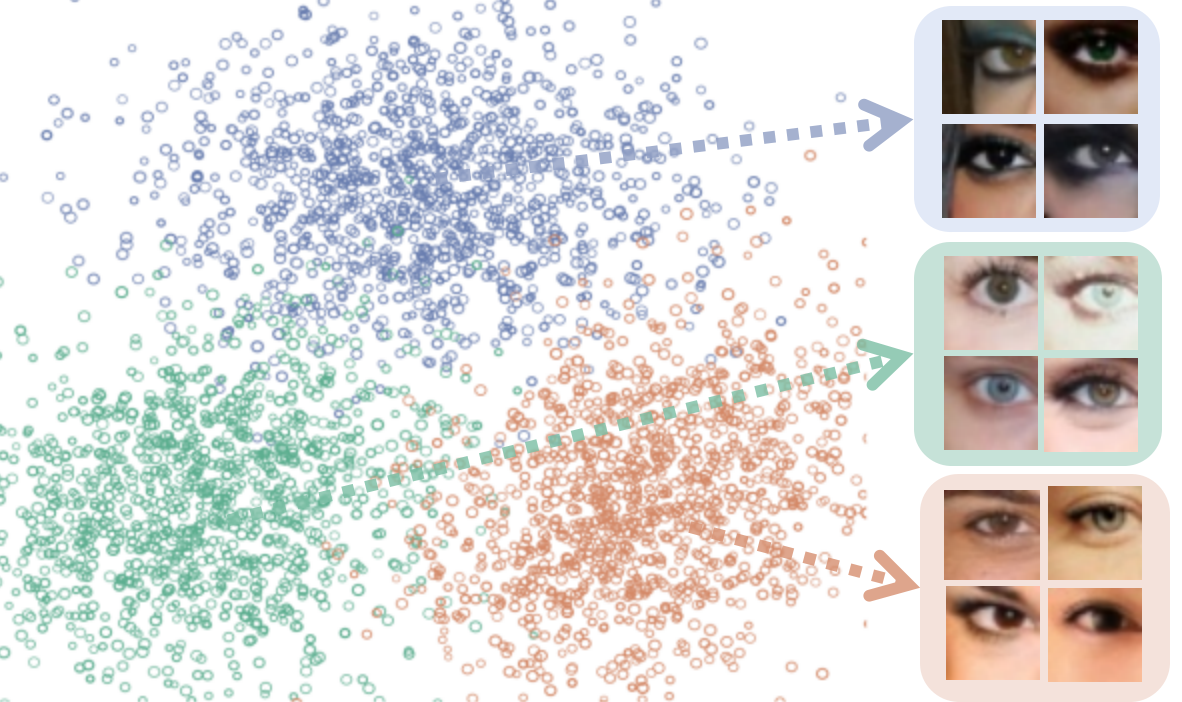}
    \caption{Feature sub-categorization.}
    \label{fig:subcate}
  \end{subfigure}
  \vspace{-6mm}
  \caption{Two key phenomena observed in facial image analysis.
  (a) After face alignment, various facial elements can be matched to corresponding positions, enhancing consistency. 
  (b) Facial features, such as eyes, are clustered into subcategories based on attributes like makeup and state, enabling finer classification.
  }
  \label{fig:motivate}
  \vspace{-6mm}
\end{figure}

To tackle this, we analyze key characteristics of facial data. As illustrated in Figure~\ref{fig:motivate}(a), faces exhibit strong spatial coherence, suggesting that masked image modeling can help capture their inherent structure. Figure~\ref{fig:motivate}(b) further shows that primary facial features, such as eyes and mouths, include fine-grained subcategories shaped by attributes like makeup and expression. Capturing these subtle variations is critical for distinguishing individuals and handling diverse facial appearances.

Based on the above insights, we present PaCo-FR: an efficient unsupervised facial representation pre-training framework that advances state-of-the-art performance through patch-pixel alignment and end-to-end codebook learning. Inspired by codebook-based models like VQ-VAE and BEiT, PaCo-FR incorporates discrete token learning into facial modeling using a novel masking and replacement strategy. Unlike conventional methods that treat image patches independently, PaCo-FR performs facial alignment to preserve spatial and structural integrity. The aligned image is divided into semantically meaningful patches, each mapped to candidate tokens from a codebook. A lightweight Belief Predictor dynamically selects tokens to replace the original patches, and the model reconstructs the masked image, learning both the semantic and geometric structure of facial features.

For high-quality training data, we curate LAION-FACE-2M-crop, a set of 2 million aligned facial images from LAION-FACE via landmark-based cropping. This ensures the model is pre-trained with spatially consistent facial data. Our approach is evaluated on face recognition, attribute prediction, and related tasks, consistently surpassing existing methods, demonstrating strong generalization and transfer abilities, even with a modest training set. These results highlight PaCo-FR's effectiveness in capturing fine-grained spatial and semantic facial characteristics, establishing a new benchmark in facial representation learning.
In summary, our main contributions are: 
\begin{itemize}
\item A new pre-training strategy that places the codebook at the decoding end, enabling end-to-end training and resolving back-propagation challenges of traditional two-stage frameworks; 
\item The introduction of a Belief Predictor to inject attribute-aware priors into token selection, improving codebook expressiveness and discrimination; 
\item End-to-end patch-level token learning for improved modeling of facial structural and semantic patterns.
\end{itemize}

\section{Related Works}
\label{sec:related_works}
\begin{figure*}
  \centering
  \includegraphics[width=0.95\linewidth]{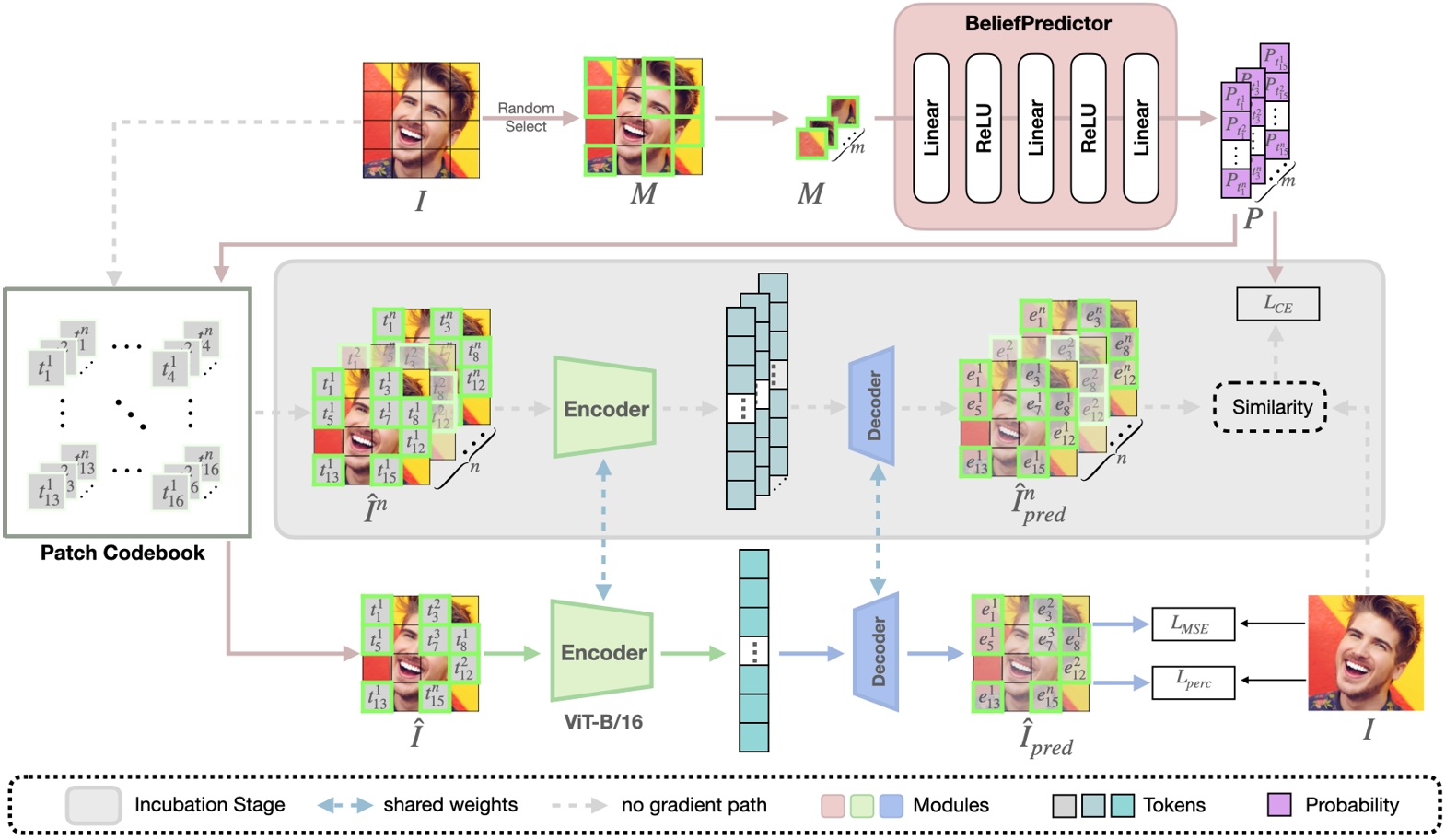}
  \vspace{-4mm} 
  \caption{The framework of PaCo-FR incorporates an incubation stage: 
  During the initial epoch of training, we supervise the predictions of the Belief Predictor, encouraging it to learn the mapping relationship from pixel space to codebook space. A part of patches($t_{*}$) are randomly selected from images, and based on the original pixel values of these patches, suitable codebook tokens($t_{*}^*$) are predicted and selected. The image $\hat{I}$, with some patches replaced, undergoes encoder and decoder processes to predict the original image $I$.}
   \label{fig:frame}
\vspace{-4mm}
\end{figure*}
\subsection{Self-Supervised Learning}

Self-supervised learning plays a pivotal role in computer vision and multi-modal pre-training.
Unlike traditional supervised methods, self-supervised learning removes reliance on manual labels by constructing meaningful image encoders that exploit the inherent semantic information within images. 
Approaches like MoCo~\cite{he2020momentum, chen2020improved, fan2021multiscale}, SimCLR~\cite{chen2020simple}, BYOL~\cite{grill2020bootstrap}, and SwAV~\cite{caron2020unsupervised} generate positive and negative samples, allowing models to learn semantic representations from large datasets. 
Methods such as CLIP~\cite{radford2021learning} and ALIGN~\cite{jia2021scaling} go beyond image-centric paradigms by leveraging supervision between images and text, enabling contrastive learning across diverse semantic spaces and aligning models' image understanding with human cognition.

\subsection{Masked Image Modeling}
The MIM paradigm is widely used in unsupervised learning, primarily involving partial occlusion of image content. 
This strategy enhances the model's understanding by framing image reconstruction as a proxy task. 
Notable methods following the MIM paradigm include MAE~\cite{he2022masked} and SimMIM~\cite{xie2022simmim}. 
Additionally, VQ-VAE~\cite{oord2017neural} and VQ-GAN~\cite{esser2021taming} demonstrate the effectiveness of a limited codebook for efficiently quantizing visual features.
Frameworks like BEiT~\cite{bao2021beit} and PeCo~\cite{dong2023peco} leverage codebooks to predict feature attributes in masked regions, enhancing the model's ability to encode image features.
However, their two-stage design results in a cumbersome training process and produces a codebook that is uninterpretable in the latent space.
The MIM approach excels in promoting the model's understanding of global spatial structure, enabling it to comprehend the content of different spatial regions within images.

\subsection{Face Pre-Training}
Training specific visual pre-training models for different tasks is a key focus of current research. 
Facial information analysis, a critical area in computer vision, still has limited pre-training methods~\cite{bulat2022pre,di2024pros}. 
FaRL~\cite{zheng2022general}, an early work in face pre-training, not only introduces a pre-training framework for facial information analysis but also proposes the LAION-FACE dataset.
Building on this, MCF~\cite{wang2023toward} presents a new face pre-training approach based on LAION-FACE, achieving superior results across multiple datasets.

\section{PaCo-FR}
\label{sec:face_clip}

\subsection{Overview}

The pipeline of PaCo-FR is shown in Figure~\ref{fig:frame}. 
The image is divided into $n$ categories, each mapped to $n$ tokens, forming a codebook space. 
Initially, $m$ patches are randomly selected from the original image $I$. 
The Belief Predictor operates on the pixel values within each patch, selecting appropriate tokens from the codebook.
This process generates a new image, $ \hat{I} $, where selected patches are replaced with tokens.
The modified image $ \hat{I} $ is input into the encoder module, which produces embeddings for all image patches. These embeddings are then passed to the decoder module, which reconstructs the predicted image $ \hat{I}_{pred} $. 
The model is trained by minimizing both the Mean Squared Error (MSE) loss and perceptual loss between the predicted image $ \hat{I}{pred} $ and the original image $I$, driving updates to both the codebook and the pre-training model.

\subsection{End-to-End Patch Codebook}

We introduce $n$ learnable tokens for each image patch, directly embedding the codebook within the image processing pipeline. 
This design effectively addresses the challenge of non-propagating gradients in traditional codebook-based methods.
Moreover, it enables the concurrent establishment of the codebook and the pre-training of facial representation learning in a single phase.

Given an original image $I$ partitioned into $K$ patches, ${I_1, \ldots, I_K}$, we randomly select $m$ patches to form the set $\mathcal{M}$. 
Each patch in $\mathcal{M}$ is fed into the Belief Predictor, which computes probabilities for various token categories. 
The most probable token is selected, replacing the corresponding patch and generating a restructured image, $\hat{I}$. The operation is defined as:
\begin{align}
    \hat{I}_i = 
\begin{cases}
I_i,  & \text{if } I_i \notin \mathcal{M},\\
t_i^{\alpha},  & \text{if } I_i \in \mathcal{M}, \alpha = \arg\max \{ p(t_{i}^j | I_i) \},
\end{cases}
\end{align}
where $i \in {1, \ldots, K}$, $j \in {1, \ldots, n}$, $t_i^j$ represents the $j$-th token corresponding to patch $I_i$, and $p(t_i^j | I_i)$ is the probability for token $t_i^j$ given patch $I_i$.

The restructured image $\hat{I}$ is then input to a standard ViT-B/16 model, which generates embeddings for each patch. 
These embeddings are passed through a decoder module consisting of multiple Transformer layers to predict the original image's pixel values. To supervise the image restoration task, we compute the MSE Loss across the entire image, similar to the MAE approach~\cite{he2022masked}. 
Additionally, to encourage the model to capture semantic information within the image, we apply a strategy inspired by PeCo~\cite{dong2023peco}. 
Both the predicted image $\hat{I}_{pred}$ and the original image $I$ are fed into a pre-trained model (kept fixed during training), calculating feature similarity across different layers of the feature maps.
The losses are computed as:
\begin{align} L_{MSE} = \parallel \hat{I}_{pred}-I \parallel_2, \end{align}
\vspace{-3mm}
\begin{align} L_{perc} = - \sum_i^m \frac{f^i(I) \centerdot f^i({\hat{I}_{pred}})}{\parallel f^i(I) \parallel_2 \cdot \parallel f^i({\hat{I}_{pred}}) \parallel_2}, \end{align}
where $m$ is the number of selected feature layers, and $f^i(*)$ represents the output of the $i$-th layer.

\subsection{Belief Predictor}

As described earlier, the Belief Predictor selects the most suitable tokens based on the content of each image patch. 
The codebook contains $n$ candidate tokens for every patch, and the model dynamically learns to map each patch to the most relevant token.
By incorporating prior knowledge, the Belief Predictor enables differential selection, ensuring that the model does not treat each patch in isolation but instead learns to prioritize and associate patches with the most meaningful representations during pre-training.

To enable the Belief Predictor to learn the mapping between pixel space and codebook space, we introduce an Incubation Stage. This stage is exclusive to the first epoch of pre-training and involves supervised training of the Belief Predictor. In this phase, we assign $n$ tokens to each patch in set $\mathcal{M}$, resulting in the creation of $\hat{I}^n$, a set containing $n$ images. All images in $\hat{I}^n$ undergo processing through both encoder and decoder modules, mapping tokens from the codebook space to the pixel space, denoted as $e$.

The similarity between the original patches $I_i$ in set $\mathcal{M}$ and their corresponding $e_i^*$ is computed, and the most similar $e_i^j$ is selected as the ground truth to supervise the Belief Predictor's output. The process is outlined as:

\begin{align}
e_i^j &= \mathcal{F}(t_i^j), \\
y_i &= \arg\min_j \left\{ \cos \langle e_i^j, I_i \rangle \right\}, \quad I_i \in \mathcal{M}, \\
L_{ce} &= -\sum_{i=1}^{m} y_i \log(P_{t_i^*}),
\end{align}
where $j \in \{1,\ldots,n\}$ and $\mathcal{F}(*)$ means outputs form encoder and decoder modules.

\section{Experiments}
\label{sec:experiments}

\subsection{Experimental Settings}
\noindent\textbf{Dataset}. 

\begin{figure}[tp]
  \centering
  \includegraphics[height=5.0cm]{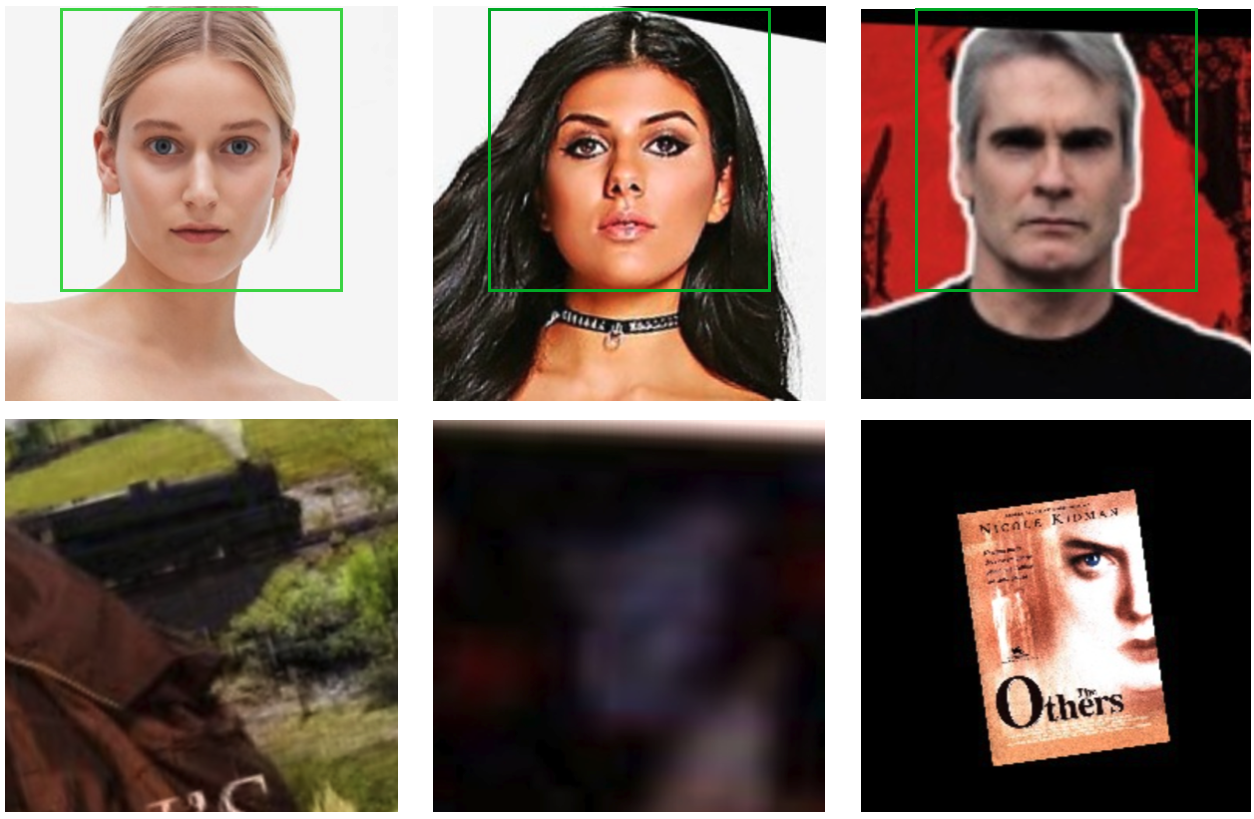}
  \vspace{-3mm}
  \caption{Cropped facial alignment results from the LAION-FACE dataset (top) and examples of noisy data present in the training set (bottom).
  }
  \label{fig:data}
\vspace{-4mm}
\end{figure}

We randomly sample 2 million images from the LAION-FACE~\cite{zheng2022general} dataset for efficient training and utilize the full dataset (approximately 20 million images) to investigate the scaling behavior of our method. Each image contains at least one human face. 
As part of the pre-processing and filtering pipeline, we randomly select and crop a single face from each image, followed by alignment to the FFHQ~\cite{karras2019stylebasedgeneratorarchitecturegenerative} standard.
The aligned faces are resized to $200 \times 200$ and padded to $256 \times 256$ with a uniform background, as shown in Figure~\ref{fig:data}.
Since LAION-FACE~\cite{zheng2022general} is constructed from the web-scale LAION dataset~\cite{schuhmann2022laion5bopenlargescaledataset} using face detection, it inevitably includes noisy samples such as non-face content, blurry faces, and low-quality images. 
Examples of such cases are illustrated in the second row of Figure~\ref{fig:data}.

\begin{table*}[t]
  \caption{
Comparison with face parsing methods and pre-trained models on the LaPa test set. 
Results are reported as F1 scores (\%, higher is better). 
\textbf{L-E}: Left Eye, \textbf{R-E}: Right Eye, \textbf{U-L}: Upper Lip, 
\textbf{I-M}: Inner Mouth, \textbf{L-L}: Lower Lip, \textbf{L-B}: Left Brow, \textbf{R-B}: Right Brow. 
\textbf{Gain} denotes the absolute improvement over the AGRNet baseline. 
The subscript $\textit{ft}$ indicates fine-tuning on LaPa.
}
\vspace{-3mm}
    \label{tab:face_parsing_lapa}
    \centering
    \setlength{\tabcolsep}{4.5pt}
    \renewcommand{\arraystretch}{0.95}
    \begin{tabular}{lccccccccccccc}
    \toprule
    \multirow{2}{*}{{Method}} & \multirow{2}{*}{{Pretrain}} & \multicolumn{10}{c}{{Facial Components}} & \multirow{2}{*}{{Mean}} & \multirow{2}{*}{{Gain}} \\
 \cmidrule(lr){3-12}
& & Skin & Hair & L-E & R-E & U-L & I-M & L-L & Nose & L-B & R-B \\\midrule
    \multicolumn{14}{c}{\textit{Non-pretrained methods}} \\
    \midrule
    {\color{gray} BASS~\cite{liu2020new}} & {\color{gray} --} & {\color{gray} 97.20} & {\color{gray} 96.30} & {\color{gray} 88.10} & {\color{gray} 88.00} & {\color{gray} 84.40} & {\color{gray} 87.60} & {\color{gray} 85.70} & {\color{gray} 95.50} & {\color{gray} 87.70} & {\color{gray} 87.60} & {\color{gray} 89.80} & {\color{gray} -2.5} \\
    {\color{gray} EHANet~\cite{luo2020ehanet}} & {\color{gray} --} & {\color{gray} 95.80} & {\color{gray} 94.30} & {\color{gray} 87.00} & {\color{gray} 89.10} & {\color{gray} 85.30} & {\color{gray} 85.60} & {\color{gray} 88.80} & {\color{gray} 94.30} & {\color{gray} 85.90} & {\color{gray} 86.10} & {\color{gray} 89.20} & {\color{gray} -3.1} \\
    {\color{gray} Wei \etal~\cite{wei2019accurate}} & {\color{gray} --} & {\color{gray} 96.10} & {\color{gray} 95.10} & {\color{gray} 88.90} & {\color{gray} 87.50} & {\color{gray} 83.10} & {\color{gray} 89.20} & {\color{gray} 83.80} & {\color{gray} 96.10} & {\color{gray} 86.00} & {\color{gray} 87.80} & {\color{gray} 89.40} & {\color{gray} -2.9} \\
    {\color{gray} EAGR~\cite{te2020edge}} & {\color{gray} --} & {\color{gray} 97.30} & {\color{gray} 96.20} & {\color{gray} 89.50} & {\color{gray} 90.00} & {\color{gray} 88.10} & {\color{gray} 90.00} & {\color{gray} 89.00} & {\color{gray} 97.10} & {\color{gray} 86.50} & {\color{gray} 87.00} & {\color{gray} 91.10} & {\color{gray} -1.2} \\
    {\color{gray} AGRNet~\cite{te2021agrnet}} & {\color{gray} --} & {\color{gray} 97.70} & {\color{gray} 96.50} & {\color{gray} 91.60} & {\color{gray} 91.10} & {\color{gray} 88.50} & {\color{gray} 90.70} & {\color{gray} 90.10} & {\color{gray} 97.30} & {\color{gray} 89.90} & {\color{gray} 90.00} & {\color{gray} 92.30} & {\color{gray} baseline} \\
    
    \midrule
    \multicolumn{14}{c}{\textit{Pre-trained methods}} \\
    \midrule
    Scratch & None & 97.06 & 92.89 & 91.43 & 91.02 & 87.06 & 89.41 & 88.94 & 97.17 & 89.71 & 89.06 & 91.37 & -0.93 \\  \midrule
    FaRL~\cite{zheng2022general} & 20M & 97.38 & 94.53 & 91.88 & 91.69 & 88.20 & 90.59 & 89.85 & 97.42 & 90.84 & 90.85 & 92.32 & +0.02 \\
    MCF~\cite{wang2023toward} & 2M & 97.30 & 94.39 & 91.67 & 91.60 & 87.86 & 90.02 & 89.70 & 97.33 & 90.44 & 89.98 & 92.03 & -0.27 \\
    \textbf{PaCo-FR} & 2M & 97.47 & 94.92 & 92.07 & 92.09 & 88.48 & 90.73 & 90.08 & 97.48 & 91.04 & 90.82 & 92.52 & \textbf{+0.22} \\
    \midrule
    FaRL\textsubscript{ft}~\cite{zheng2022general} & 20M & 97.52 & 95.11 & 92.33 & 92.09 & 88.69 & 90.70 & 90.05 & 97.55 & 91.57 & 91.34 & 92.70 & +0.40 \\
    MCF\textsubscript{ft}~\cite{wang2023toward} & 2M & 97.46 & 94.93 & 92.26 & 92.22 & 88.31 & 90.20 & 89.84 & 97.53 & 91.20 & 90.79 & 92.48 & +0.18 \\
    \textbf{PaCo-FR\textsubscript{ft}} & 2M & \textbf{97.63} & \textbf{95.43} & \textbf{92.32} & \textbf{92.43} & \textbf{88.99} & \textbf{90.82} & \textbf{90.21} & \textbf{97.69} & \textbf{91.74} & \textbf{91.27} & \textbf{92.85} & \textbf{+0.55} \\
    \bottomrule
    \end{tabular}
    \vspace{-5mm}
\end{table*}

\begin{table*}[tb]
    \centering
    \setlength{\tabcolsep}{3.8pt}
    \renewcommand{\arraystretch}{0.95}
    \caption{
Comparison with face parsing methods and pre-trained models on the CelebAMask-HQ test set. 
Results are reported as F1 scores (\%, higher is better). 
\textbf{L/R-Eye}: Left/Right Eye, \textbf{L/R-Brow}: Left/Right Eyebrow, \textbf{L/R-Ear}: Left/Right Ear, 
\textbf{I-Mouth}: Inner Mouth, \textbf{U/L-Lip}: Upper/Lower Lip. 
\textbf{Gain} denotes the absolute improvement over the AGRNet baseline. 
The subscript $_\textit{ft}$ indicates fine-tuning on CelebAMask-HQ.
}
 \vspace{-3mm}
   \label{tab:face_parsing_celebm}
    
    \begin{tabular}{lcccccccccccc}
    \toprule
\multirow{3}{*}{{Method}} & \multirow{3}{*}{{Pretrain}} & \multicolumn{9}{c}{Facial Components} & \multirow{3}{*}{{Mean}} & \multirow{3}{*}{{Gain}} \\
\cmidrule(lr){3-11}
& & Face & Nose & Glasses & L-Eye & R-Eye & L-Brow & R-Brow & L-Ear & R-Ear \\
\cmidrule(lr){3-11}
& & I-Mouth & U-Lip & L-Lip & Hair & Hat & Earring & Necklace & Neck & Cloth \\
\midrule
    \multicolumn{13}{c}{\textit{Non-pretrained methods}} \\
\midrule
& & \textcolor{gray}{63.4} & \textcolor{gray}{88.9} & \textcolor{gray}{90.1} & \textcolor{gray}{86.6} & \textcolor{gray}{91.3} & \textcolor{gray}{63.2} & \textcolor{gray}{26.1} & \textcolor{gray}{92.8} & \textcolor{gray}{68.3} & & \\
\textcolor{gray}{\multirow{2}{*}{EHANet~\cite{luo2020ehanet}}} & \textcolor{gray}{--} & \textcolor{gray}{96.0} & \textcolor{gray}{93.7} & \textcolor{gray}{90.6} & \textcolor{gray}{86.2} & \textcolor{gray}{86.5} & \textcolor{gray}{83.2} & \textcolor{gray}{83.1} & \textcolor{gray}{86.5} & \textcolor{gray}{84.1} & \multirow{2}{*}{\textcolor{gray}{84.0}} & \multirow{2}{*}{\textcolor{gray}{-1.5}} \\
& & \textcolor{gray}{93.8} & \textcolor{gray}{88.6} & \textcolor{gray}{90.3} & \textcolor{gray}{93.9} & \textcolor{gray}{85.9} & \textcolor{gray}{67.8} & \textcolor{gray}{30.1} & \textcolor{gray}{88.8} & \textcolor{gray}{83.5} & & \\
\textcolor{gray}{\multirow{2}{*}{Wei \etal~\cite{wei2019accurate}}} & \textcolor{gray}{--} & \textcolor{gray}{96.4} & \textcolor{gray}{91.9} & \textcolor{gray}{89.5} & \textcolor{gray}{87.1} & \textcolor{gray}{85.0} & \textcolor{gray}{80.8} & \textcolor{gray}{82.5} & \textcolor{gray}{84.1} & \textcolor{gray}{83.3} & \multirow{2}{*}{\textcolor{gray}{82.1}} & \multirow{2}{*}{\textcolor{gray}{-3.4}} \\
& & \textcolor{gray}{90.6} & \textcolor{gray}{87.9} & \textcolor{gray}{91.0} & \textcolor{gray}{91.1} & \textcolor{gray}{83.9} & \textcolor{gray}{65.4} & \textcolor{gray}{17.8} & \textcolor{gray}{88.1} & \textcolor{gray}{80.6} & & \\
\textcolor{gray}{\multirow{2}{*}{EAGR~\cite{te2020edge}}} & \textcolor{gray}{--} & \textcolor{gray}{96.2} & \textcolor{gray}{94.0} & \textcolor{gray}{92.3} & \textcolor{gray}{88.6} & \textcolor{gray}{88.7} & \textcolor{gray}{85.7} & \textcolor{gray}{85.2} & \textcolor{gray}{88.0} & \textcolor{gray}{85.7} & \multirow{2}{*}{\textcolor{gray}{85.1}} & \multirow{2}{*}{\textcolor{gray}{-0.4}} \\
& & \textcolor{gray}{95.0} & \textcolor{gray}{88.9} & \textcolor{gray}{91.2} & \textcolor{gray}{94.9} & \textcolor{gray}{87.6} & \textcolor{gray}{68.3} & \textcolor{gray}{27.6} & \textcolor{gray}{89.4} & \textcolor{gray}{85.3} & & \\
\textcolor{gray}{\multirow{2}{*}{AGRNet~\cite{te2021agrnet}}} & \textcolor{gray}{--} & \textcolor{gray}{96.5} & \textcolor{gray}{93.9} & \textcolor{gray}{91.8} & \textcolor{gray}{88.7} & \textcolor{gray}{89.1} & \textcolor{gray}{85.5} & \textcolor{gray}{85.6} & \textcolor{gray}{88.1} & \textcolor{gray}{88.7} & \multirow{2}{*}{\textcolor{gray}{85.5}} & \multirow{2}{*}{\textcolor{gray}{baseline}} \\
& & \textcolor{gray}{92.0} & \textcolor{gray}{89.1} & \textcolor{gray}{91.1} & \textcolor{gray}{95.2} & \textcolor{gray}{87.2} & \textcolor{gray}{69.6} & \textcolor{gray}{32.8} & \textcolor{gray}{89.9} & \textcolor{gray}{84.9} & & \\
\midrule
    \multicolumn{13}{c}{\textit{Pre-trained methods}} \\
    \midrule
\multirow{2}{*}{Scratch} & \multirow{2}{*}{None} & 96.18 & 93.72 & 92.21 & 88.80 & 88.70 & 85.09 & 85.25 & 86.68 & 87.25 & \multirow{2}{*}{85.05} & \multirow{2}{*}{-0.45} \\
& & 90.88 & 87.51 & 89.76 & 95.11 & 85.20 & 62.93 & 37.81 & 91.03 & 86.73 & & \\
\midrule
\multirow{2}{*}{FaRL~\cite{zheng2022general}} & \multirow{2}{*}{20M} & 96.29 & 93.72 & 93.91 & 88.75 & 88.64 & 85.24 & 85.42 & 87.06 & 87.36 & \multirow{2}{*}{86.72} & \multirow{2}{*}{+1.22} \\
& & 90.96 & 87.53 & 89.81 & 95.60 & 90.07 & 68.19 & 50.94 & 91.54 & 89.88 & & \\

\multirow{2}{*}{MCF~\cite{wang2023toward}} & \multirow{2}{*}{2M} & 96.33 & 93.85 & 93.48 & 88.94 & 88.80 & 85.40 & 85.58 & 87.16 & 87.63 & \multirow{2}{*}{86.09} & \multirow{2}{*}{+0.59} \\
& & 91.17 & 87.86 & 89.98 & 95.47 & 88.22 & 65.65 & 44.04 & 91.35 & 88.77 & & \\

\multirow{2}{*}{\textbf{PaCo-FR}} & \multirow{2}{*}{2M} & 96.45 & 93.99 & 94.14 & 89.37 & 89.15 & 85.90 & 86.04 & 87.78 & 88.13 & \multirow{2}{*}{87.33} & \multirow{2}{*}{+1.83} \\

& & 91.75 & 88.63 & 90.45 & 95.63 & 88.56 & 69.06 & 55.85 & 91.76 & 89.28 & & \\
\midrule
\multirow{2}{*}{FaRL\textsubscript{ft}~\cite{zheng2022general}} & \multirow{2}{*}{20M} & 96.32 & 93.62 & 94.08 & 88.81 & 88.67 & 85.25 & 85.46 & 87.53 & 87.87 & \multirow{2}{*}{87.55} & \multirow{2}{*}{+2.05} \\
& & 91.10 & 87.77 & 89.81 & 95.76 & 90.80 & 69.87 & 60.91 & 91.79 & 90.40 & & \\

\multirow{2}{*}{MCF\textsubscript{ft}~\cite{wang2023toward}} & \multirow{2}{*}{2M} & 96.38 & 93.86 & 93.43 & 89.00 & 88.90 & 85.58 & 85.73 & 87.47 & 87.98 & \multirow{2}{*}{86.69} & \multirow{2}{*}{+1.19} \\
& & 91.33 & 87.98 & 90.12 & 95.69 & 89.91 & 67.28 & 48.68 & 91.65 & 89.44 & & \\

\multirow{2}{*}{\textbf{PaCo-FR\textsubscript{ft}}} & \multirow{2}{*}{2M} 
& \textbf{96.47} & \textbf{94.03} & \textbf{94.49} 
& \textbf{89.37} & \textbf{89.27} & \textbf{86.01} 
& \textbf{86.06} & 87.80 & 88.33 
& \multirow{2}{*}{\textbf{87.77}} & \multirow{2}{*}{\textbf{+2.27}} \\
& & \textbf{91.91} & 88.62 & \textbf{90.50} 
& \textbf{95.81} & \textbf{90.32} & \textbf{70.07} 
& \textbf{58.66} & \textbf{91.94} & \textbf{90.24} & &\\

    \bottomrule
    \end{tabular}
    \vspace{-6mm}
\end{table*}

\begin{table}[tb]
  \caption{
  Comparison with previous face alignment methods and face pre-trained models on the 300W test set, evaluated by Normalized Mean Error (NME, \%) with inter-ocular normalization. Results are shown for the Common subset, Challenge subset, and Full set.
  }
  \label{tab:face_landmark_300w}
  \centering
  \vspace{-3mm}
 \begin{adjustbox}{width=\columnwidth}
  \begin{tabular}{l c ccc}
   \toprule
    \multirow{2}{*}{{Method}} & \multirow{2}{*}{{Pretrain}} & \multicolumn{3}{c}{{NME$_{\text{inter-ocular}}$} $\downarrow$} \\
    \cmidrule(lr){3-5}
    & & Common & Challenge & Full \\
    \midrule
    \multicolumn{5}{c}{\textit{{Previous methods}}} \\ \midrule
    \textcolor{gray}{LAB (w/ B)~\cite{wu2018look}} & \textcolor{gray}{--} & \textcolor{gray}{2.98} & \textcolor{gray}{5.19} & \textcolor{gray}{3.49} \\
    \textcolor{gray}{DCFE (w/ 3D)~\cite{valle2018deeply}} & \textcolor{gray}{--} & \textcolor{gray}{2.76} & \textcolor{gray}{5.22} & \textcolor{gray}{3.24} \\
    \textcolor{gray}{DeCaFa~\cite{dapogny2019decafa}} & \textcolor{gray}{--} & \textcolor{gray}{2.93} & \textcolor{gray}{5.26} & \textcolor{gray}{3.39} \\
    \textcolor{gray}{HR-Net~\cite{sun2019high}} & \textcolor{gray}{--} & \textcolor{gray}{2.87} & \textcolor{gray}{5.15} & \textcolor{gray}{3.32} \\
    \textcolor{gray}{HG-HSLE~\cite{zou2019learning}} & \textcolor{gray}{--} & \textcolor{gray}{2.85} & \textcolor{gray}{5.03} & \textcolor{gray}{3.28} \\
    \textcolor{gray}{AWing~\cite{wang2019adaptive}} & \textcolor{gray}{--} & \textcolor{gray}{2.72} & \textcolor{gray}{4.52} & \textcolor{gray}{3.07} \\
    \textcolor{gray}{LUVLi~\cite{kumar2020luvli}} & \textcolor{gray}{--} & \textcolor{gray}{2.76} & \textcolor{gray}{5.16} & \textcolor{gray}{3.23} \\
    \textcolor{gray}{ADNet~\cite{huang2021adnet}} & \textcolor{gray}{--} & \textcolor{gray}{\textbf{2.53}} & \textcolor{gray}{4.58} & \textcolor{gray}{\textbf{2.93}} \\
    \midrule
    \multicolumn{5}{c}{\textit{Pre-trained methods}} \\ \midrule
    Scratch & None & 2.89 & 5.09 & 3.32 \\\midrule
    FaRL~\cite{zheng2022general} & 20M & 2.69 & 4.85 & 3.12 \\
    MCF~\cite{wang2023toward} & 2M & 2.67 & 4.74 & 3.07 \\
    \textbf{PaCo-FR} & 2M & \underline{2.60} & 4.65 & \underline{3.00} \\
    \midrule
    FaRL\textsubscript{ft}~\cite{zheng2022general} & 20M & 2.70 & \textbf{4.64} & 3.08 \\
    MCF\textsubscript{ft}~\cite{wang2023toward} & 2M & 2.65 & \textbf{4.50} & 3.06 \\
    \textbf{PaCo-FR\textsubscript{ft}} & 2M & 2.62 & \textbf{4.50} & \underline{3.00} \\
    \bottomrule
  \end{tabular}
  
  \end{adjustbox}
\vspace{-6mm}
\end{table}

\begin{table}[tb]
  \caption{
  Comparison with previous face alignment methods and pre-trained models on the AFLW-19 test sets. 
  Results are reported as Normalized Mean Error (NME, \%) with face diagonal and bounding box normalization. 
  AUC${}_\text{box}^7$ represents the Area Under Curve at 7\% threshold. 
  }
  \label{tab:face_landmark_aflw-19}
  \centering
  \vspace{-3mm}
  \begin{adjustbox}{width=\columnwidth}
  \begin{tabular}{l c cc cc}
    \toprule
    \multirow{2}{*}{{Method}} & \multirow{2}{*}{Pretrain} & \multicolumn{2}{c}{NME$_\text{diag}$ $\downarrow$} & NME$_\text{box}$ $\downarrow$ & AUC$_\text{box}^7$ $\uparrow$ \\
    \cmidrule(lr){3-4} \cmidrule(lr){5-6}
    &  & Full & Frontal & Full & Full \\
    \midrule
    \multicolumn{6}{c}{\textit{{Previous methods}}} \\ \midrule
    \textcolor{gray}{LAB (w/ B)~\cite{wu2018look}} & \textcolor{gray}{--} & \textcolor{gray}{1.25} & \textcolor{gray}{1.14} & \textcolor{gray}{--} & \textcolor{gray}{--} \\
    \textcolor{gray}{HR-Net~\cite{sun2019high}} & \textcolor{gray}{--} & \textcolor{gray}{1.57} & \textcolor{gray}{1.46} & \textcolor{gray}{--} & \textcolor{gray}{--} \\
    \textcolor{gray}{Wing~\cite{feng2018wing}} & \textcolor{gray}{--} & \textcolor{gray}{--} & \textcolor{gray}{--} & \textcolor{gray}{3.56} & \textcolor{gray}{53.5} \\
    \textcolor{gray}{KDN~\cite{chen2019face}} & \textcolor{gray}{--} & \textcolor{gray}{--} & \textcolor{gray}{--} & \textcolor{gray}{2.80} & \textcolor{gray}{60.3} \\
    \textcolor{gray}{LUVLi~\cite{kumar2020luvli}} & \textcolor{gray}{--} & \textcolor{gray}{1.39} & \textcolor{gray}{1.19} & \textcolor{gray}{2.28} & \textcolor{gray}{68.0} \\
    \midrule
    \multicolumn{6}{c}{\textit{Pre-trained methods}} \\\midrule
    Scratch & None & 1.035 & 0.868 & 1.463 & 79.5 \\\midrule
    FaRL~\cite{zheng2022general} & 20M & 0.991 & 0.851 & 1.402 & 80.4 \\
    MCF~\cite{wang2023toward} & 2M & 0.999 & 0.851 & 1.414 & 80.2 \\
    \textbf{PaCo-FR} & 2M & 0.990 & 0.846 & 1.400 & 80.4 \\
    \midrule
    FaRL\textsubscript{ft}~\cite{zheng2022general} & 20M & 0.969 & 0.836 & 1.371 & \underline{80.8} \\
    MCF\textsubscript{ft}~\cite{wang2023toward} & 2M & \underline{0.968} & \underline{0.827} & \underline{1.368} & \underline{80.8} \\
    \textbf{PaCo-FR\textsubscript{ft}} & 2M & \textbf{0.955} & \textbf{0.820} & \textbf{1.351} & \textbf{81.1} \\
    \bottomrule
  \end{tabular}
  \end{adjustbox}
  \vspace{-5mm}
\end{table}

\begin{table}[tb]
  \caption{
  Comparison with previous face alignment methods and pre-trained models on WFLW test sets. 
  Lower values ($\downarrow$) indicate better performance. 
  \textbf{Expr.}: Expression. \textbf{Illum.}: Illumination. \textbf{Occl.}: Occlusion.
  }  \vspace{-3mm}
  \label{tab:face_landmark_wflw}
  \centering
  \setlength{\tabcolsep}{3.8pt}
  \renewcommand{\arraystretch}{0.95}
  \begin{adjustbox}{width=\columnwidth}
  \begin{tabular}{lcccccccc}
    \toprule
    \multirow{2}{*}{{Method}} & \multirow{2}{*}{{Pre-train}} & \multicolumn{7}{c}{{$\mathrm{NME}_\mathrm{inter\textit{-}ocular}$ $\downarrow$}} \\
    \cmidrule{3-9}
    &  & {Full} & {Pose} & {Expr.} & {Illum.} & {MakeUp} & {Occl.} & {Blur} \\
    \midrule
    \multicolumn{9}{c}{\textit{{Previous methods}}} \\
    \midrule
    \textcolor{gray}{LAB~\cite{wu2018look}} & \textcolor{gray}{--} & \textcolor{gray}{5.27} & \textcolor{gray}{10.24} & \textcolor{gray}{5.51} & \textcolor{gray}{5.23} & \textcolor{gray}{5.15} & \textcolor{gray}{6.79} & \textcolor{gray}{6.12} \\
    \textcolor{gray}{Wing~\cite{feng2018wing}} & \textcolor{gray}{--} & \textcolor{gray}{5.11} & \textcolor{gray}{8.75} & \textcolor{gray}{5.36} & \textcolor{gray}{4.93} & \textcolor{gray}{5.41} & \textcolor{gray}{6.37} & \textcolor{gray}{5.81} \\
    \textcolor{gray}{DeCaFA~\cite{dapogny2019decafa}} & \textcolor{gray}{--} & \textcolor{gray}{4.62} & \textcolor{gray}{8.11} & \textcolor{gray}{4.65} & \textcolor{gray}{4.41} & \textcolor{gray}{4.63} & \textcolor{gray}{5.74} & \textcolor{gray}{5.38} \\
    \textcolor{gray}{AWing~\cite{wang2019adaptive}} & \textcolor{gray}{--} & \textcolor{gray}{4.36} & \textcolor{gray}{7.38} & \textcolor{gray}{4.58} & \textcolor{gray}{4.32} & \textcolor{gray}{4.27} & \textcolor{gray}{5.19} & \textcolor{gray}{4.96} \\
    \textcolor{gray}{ADNet~\cite{huang2021adnet}} & \textcolor{gray}{--} & \textcolor{gray}{\underline{4.14}} & \textcolor{gray}{6.96} & \textcolor{gray}{4.38} & \textcolor{gray}{4.09} & \textcolor{gray}{4.05} & \textcolor{gray}{5.06} & \textcolor{gray}{4.79} \\
    \midrule
    \multicolumn{9}{c}{\textit{Pre-trained methods}} \\
    \midrule
    Scratch & None & 4.95 & 8.95 & 5.22 & 4.84 & 5.19 & 6.31 & 5.46 \\
    FaRL~\cite{zheng2022general} & 20M & 4.38 & 7.60 & 4.66 & 4.19 & 4.30 & 5.44 & 4.98 \\
    MCF~\cite{wang2023toward} & 2M & 4.51 & 7.93 & 4.73 & 4.44 & 4.53 & 5.63 & 5.17 \\
    PaCo-FR & 2M & 4.35 & 7.50 & 4.52 & 4.22 & 4.18 & 5.43 & 5.02 \\
    FaRL\textsubscript{ft}~\cite{zheng2022general} & 20M & 4.03 & \underline{6.81} & \underline{4.32} & \underline{3.92} & \underline{3.87} & \textbf{4.70} & \underline{4.54} \\
    MCF\textsubscript{ft}~\cite{wang2023toward} & 2M & 4.16 & 7.15 & 4.40 & 4.05 & 4.09 & 4.99 & 4.66 \\
    PaCo-FR\textsubscript{ft} & 2M & \textbf{3.99} & \textbf{6.75} & \textbf{4.21} & \textbf{3.87} & \textbf{3.80} & \underline{4.71} & \textbf{4.47} \\
    \bottomrule
  \end{tabular}
  \end{adjustbox}
  \vspace{-4mm}
\end{table}

\noindent\textbf{Pre-training Setup}. We adopt a standard ViT-B/16 architecture~\cite{dosovitskiy2021imageworth16x16words} as the encoder, operating on $224 \times 224$ input images. 
The decoder consists of a single Transformer layer followed by a linear projection layer. Following MAE~\cite{he2022masked}, we construct the masked patch set $\mathcal{M}$ by randomly sampling 75\% of all patches.
The codebook space is defined with a final dimension of $3 \times 196$, meaning each of the 196 patches is associated with 3 candidate tokens.
We employ a MoCo-v3~\cite{chen2021empirical} pre-trained model to compute the perceptual loss.
During the \textit{Incubation Stage}, dropout is applied to the Belief Predictor to prevent training collapse. 
Specifically, 75\% of the patches in $\mathcal{M}$ are replaced with random tokens, while the remaining patches are filled with tokens selected by the Belief Predictor.
The predictions are supervised to ensure training stability and accurate token learning.
We use a batch size of 720 and set the weight decay to 0.05. The model is trained for 32 epochs on the LAION-FACE-2M dataset, with 1 epoch of linear warmup. 
All experiments are conducted using 8 NVIDIA Tesla V100 GPUs, and the total pre-training time is approximately 17 hours.

\noindent\textbf{Implementation Details}. For downstream tasks, we use our pretrained encoder as the backbone network. We design specialized task heads for different task categories. To enable multi-level facial feature extraction, we feed outputs from layers {2, 4, 8, 12} of the encoder to the task heads.

\subsection{2D Facial Analysis Tasks}
\label{ssec:sota}
We employ our pretrained model as a backbone network for various 2D facial analysis tasks, which collectively assess five key capabilities: local feature discrimination, spatial relationship modeling, perturbation invariance, fine-grained representation learning, and cross-domain generalization. To comprehensively evaluate these aspects, we conduct experiments on two fundamental tasks - face parsing and face alignment - using two and three distinct benchmark datasets respectively for each task.

\noindent\textbf{Face Parsing}. The face parsing task involves decomposing a facial image into different semantic parts or regions, typically identifying and segmenting various components of the face such as eyes, mouth, nose, etc. This task aims to understand the low-level semantic information of faces in detail, providing finer facial representations for computer vision applications. Evaluating our model on the face parsing task allows us to assess its capabilities in understanding facial semantics, distinguishing facial color, and texture features. We selected the LaPa~\cite{liu2020new} and CelebAMask-HQ~\cite{lee2020maskgan} datasets from the face parsing task to compare the performance of our model, other dedicated models, and facial pre-training models. The LaPa dataset includes over 22K images, with 18176 images in the training set and 2K images in the test set, each containing 11-category pixel-level label map. The CelebAMask-HQ dataset contains approximately 30K high-quality facial images. The training set includes 24813 images, and the test set comprises 2824 images, each containing 19-category pixel-level label map. 
Following~\cite{liu2020new,te2020edge,te2021agrnet},  
the F1 scores of facial components are used to measure the performance.

Tables~\ref{tab:face_parsing_lapa} and \ref{tab:face_parsing_celebm} respectively show the performance of our method compared to other specific face parsing task models and face pre-trained models on the LaPa and CelebAMask-HQ datasets. From the tables, it can be observed that when the image size is 224, our method, under both freeze and refine settings for the backbone, outperforms FaRL~\cite{zheng2022general} pretrained on 20 million data samples and MCF~\cite{wang2023toward} pretrained on an equivalent 2 million data samples. However, in the Celebm dataset, when the image is scaled to 448, F1 score of our model for the category ``Hat'' is approximately 2\% lower than FaRL, resulting in our average F1 score being 0.1\% lower than FaRL in this scenario. Our model achieves the highest F1 scores on the LaPa dataset, reaching 93.91\%.

\begin{table}[tb]
  \caption{Performance comparison of PaCo-FR models across facial analysis tasks with varying training data sizes and input resolutions. 
LaPaM.: \textbf{LaPa}$_{\text{Mean}}$$\uparrow$, CA.: \textbf{CelebAMask-HQ}$_{\text{Mean}}$$\uparrow$, 300W.: \textbf{300W}$_{\text{NME}}$$\downarrow$, AFLW.: \textbf{AFLW-19}$_{\text{NME}}$$\downarrow$, and WFL.: \textbf{WFLW}$_{\text{NME}}$$\downarrow$.
  ${\textit{ft}}$ denotes fine-tuned models. ${448}$ indicates models trained with 448x448 pixel input resolution (vs. standard 224x224 pixel input).
  }
  \vspace{-3mm}
  \label{tab:scale}
  \centering
  \setlength{\tabcolsep}{3.8pt}
  \begin{adjustbox}{width=\columnwidth}
  \footnotesize
  \renewcommand{\arraystretch}{0.95}
  \begin{tabular}{lcccccc}
    \toprule
    \multirow{2}{*}{{Method}} & \multirow{2}{*}{{Pre-train}} & \multicolumn{5}{c}{{Evaluation Metrics}} \\
    \cmidrule{3-7}
    && {LPaM.} & {CA.} & {300W.} & {AFLW.}& {WFL.} \\
    \midrule
     PaCo-FR & 2M & 92.52 & 87.33 & 3.00 & 0.99 & 4.35 \\
     PaCo-FR$_{\textit{ft}}$ & 2M & 92.85 & 87.33 & 3.00 & 0.955 & 3.99 \\
    \rowcolor{green!15!gray!10} PaCo-FR & 20M & 92.58 & 87.58 & 2.99 & 0.991 & 4.45 \\
    \rowcolor{green!15!gray!10} PaCo-FR$_{\textit{ft}}$ & 20M & 93.05 & 88.05 & 2.92 & 0.951 & 3.88 \\
    \hdashline
     PaCo-FR$^{448}$ & 2M & 93.43 & 89.46 & 2.89 & 0.940 & 3.93 \\
    \rowcolor{green!15!gray!10} PaCo-FR$_{\textit{ft}}^{448}$ & 20M & \textbf{93.91} & \textbf{89.63} & \textbf{2.83} & \textbf{0.932} & \textbf{3.86} \\
    \bottomrule
  \end{tabular}
  \end{adjustbox}
  \vspace{-4mm}
\end{table}

\noindent\textbf{Face Alignment}. The face alignment task focuses on the precise localization of facial keypoints, enabling the model to better understand the structure and shape of faces. This understanding provides advantageous input for subsequent facial analysis tasks. Evaluating the model on such tasks allows us to assess its capability to comprehend overall facial feature information, thereby enriching tasks like face recognition and emotion analysis with more detailed insights. We have selected three different face alignment datasets, namely 300W~\cite{sagonas2016300,sagonas2013semi,sagonas2013300}, AFLW-19~\cite{zhu2016unconstrained}, and WFLW~\cite{wu2018look}. The 300W dataset includes 3837 images for training and 600 images for testing, each containing 68 landmarks. The AFLW-19 dataset consists of 20K images for training and 4386 images for testing, each with 19 landmarks. The WFLW dataset includes 7500 images for training and 2500 images for testing, each with 98 landmarks. Following common practice, we use normalized mean error (NME), failure rate (FR) and AUC as the metric.

Tables~\ref{tab:face_landmark_300w}, ~\ref{tab:face_landmark_aflw-19}, and ~\ref{tab:face_landmark_wflw} showcase the performance of our method on three datasets: 300W, AFLW-19, and WFLW. Our method, under three different settings, achieves overall scores surpassing FaRL~\cite{zheng2022general} pretrained on 20 million data samples and MCF~\cite{wang2023toward} pretrained on an equivalent 2 million data samples. Furthermore, the final accuracy exceeds that of other specific face parsing task models.

\noindent\textbf{Scaling Law}. Our method demonstrates superior efficiency, achieving better performance with only 2M samples than FaRL trained on 10x larger datasets. To establish scaling laws for our approach, we scaled the data size to 20M samples. Additionally, we investigated model behavior by scaling input dimensions. As shown in Table~\ref{tab:scale}, our method achieves state-of-the-art results in the final configuration.

\begin{figure*}[tb]
  \centering
  \includegraphics[width=\linewidth]{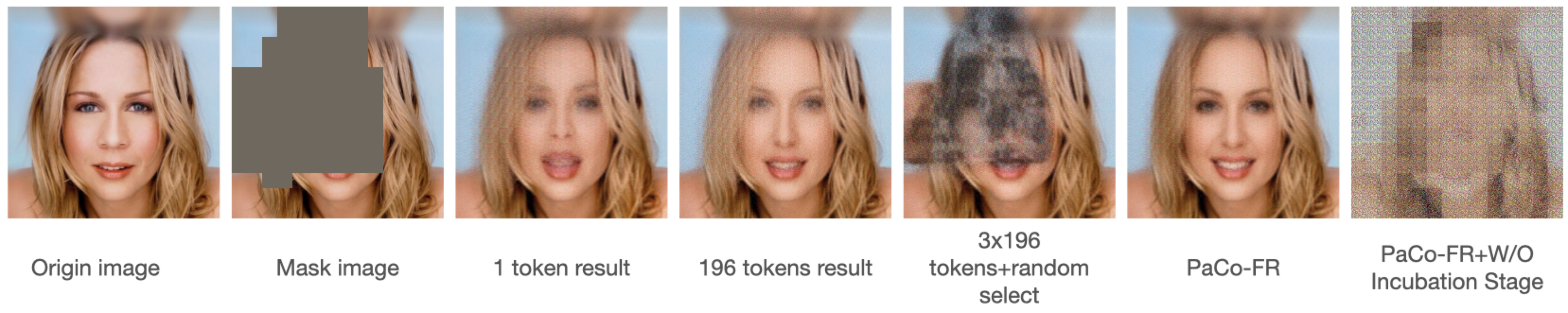}
  \vspace{-8mm}
   \caption{The visualizations depict the impact of codebook size and different configurations on the generative capabilities of the model in the proposed method.
   }
   \label{fig:ablation}
  \vspace{-5mm}
\end{figure*}

\begin{table}[tb]
  \caption{Ablation Study Results for PaCo-FR and Comparative Analysis of Codebook Space Size Variations. 
  The results include the performance on LaPa (F1-mean $\uparrow$) and AFLW-19 (Normalized Mean Error $\downarrow$).}
  \vspace{-3mm}
  \label{tab:ablation}
  \centering
  \begin{adjustbox}{width=\columnwidth}
  \begin{tabular}{lcc}
  \toprule
  {Pre-training Settings} & {LaPa}$_{\text{F1-mean} \uparrow}$ & {AFLW-19}$_{\text{NME}_\text{diag} \downarrow}$ \\
  \midrule
   \multicolumn{3}{c}{\textit{Codebook token configuration}} \\
  \hdashline
   1 token (MAE method) & 92.12 & 1.021 \\
   $1 \times 196$ tokens & \underline{92.30} & \underline{0.991} \\
   $3 \times 196$ tokens + random selection & 92.24 & 1.011 \\
   $5 \times 196$ tokens + random selection & 92.07 & 1.103 \\
   \rowcolor{green!15!gray!10}\emph{$3 \times 196$} tokens + Belief Predictor (PaCo-FR) & \textbf{92.52} & \textbf{0.990} \\
   $5 \times 196$ tokens + Belief Predictor (PaCo-FR) & 92.42 & 0.993 \\
  \hdashline
   \multicolumn{3}{c}{\textit{Training strategy}} \\
  \hdashline
   PaCo-FR + W/O Incubation Stage & 88.71 & 1.310 \\
   \rowcolor{green!15!gray!10} PaCo-FR + W Incubation Stage (PaCo-FR) & \textbf{92.52} & \textbf{0.990} \\
  \bottomrule
  \end{tabular}
  \end{adjustbox}
  
  \vspace{-0.3mm}
\end{table}

\subsection{3D Face Reconstruction}
Our 3D face reconstruction framework extends the MICA\cite{zielonka2022mica} architecture, which employs an ArcFace\cite{deng2019arcface} backbone to extract facial identity features and an MLP mapping network to regress shape parameters. While MICA effectively reconstructs neutral 3D facial geometry from single images, it lacks the capability to model facial expressions. To address this limitation, we augment the architecture by introducing a parallel expression prediction branch that leverages pretrained facial representation encoders such as PaCo-FR. The complete framework (illustrated in supplementary document) is trained on the CoMA\cite{COMA:ECCV18} and FaMoS\cite{TMPEH:CVPR:2023} datasets, where FLAME\cite{FLAME:SiggraphAsia2017} parameter fitting provides disentangled shape and expression representations. This extension preserves MICA's shape reconstruction accuracy while enabling expressive face modeling through an end-to-end trainable architecture with multi-task objectives.

\begin{figure}[tp]
  \centering
  \includegraphics[height=6.2cm]{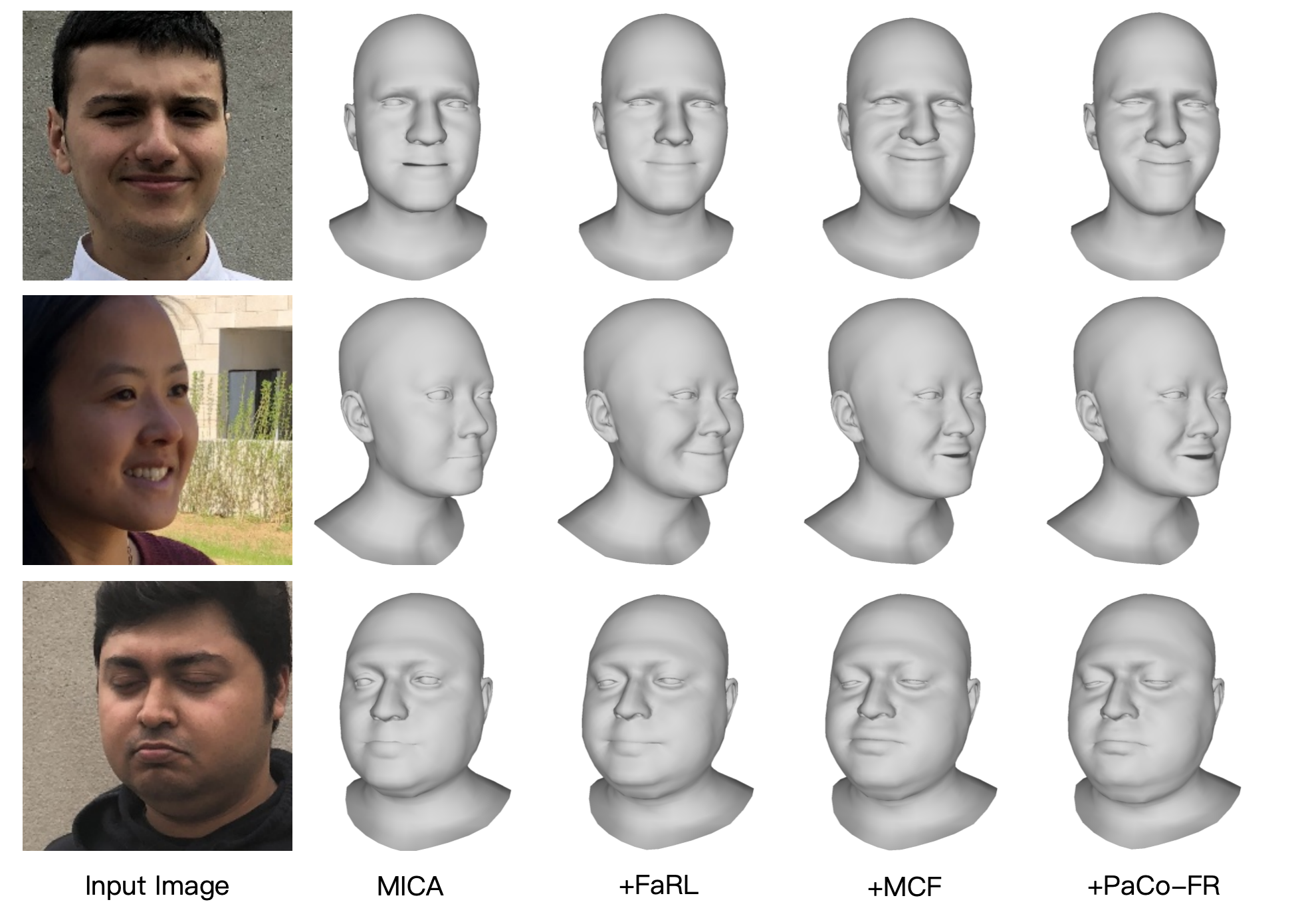}
  \vspace{-8mm}
  \caption{Comparison of expression reconstruction on NoW validation images.
  }
  \label{fig:recon_comp}
\vspace{-5mm}
\end{figure}

To evaluate the accuracy of 3D face reconstruction, we adopt the Now\cite{RingNet:CVPR:2019} Validation benchmark, which reports the mean, median, and standard deviation of mean square error (MSE) across two sub-benchmarks: Non-Metrical and Metrical. We conduct comparative experiments between the original MICA framework (without expression prediction) and our extended version incorporating different pretrained facial representation encoders for expression estimation. As quantitatively demonstrated in Table~\ref{tab:face_reconstruction_now}, our approach with PaCo-FR as the expression branch backbone achieves superior performance, yielding the lowest MSE scores on both the Non-Metrical and Metrical benchmarks.

Figure~\ref{fig:recon_comp} presents a comparative visualization of expression reconstruction performance across different methods. The results demonstrate that while MICA achieves satisfactory geometric accuracy in facial shape reconstruction, its output is inherently limited to neutral expressions due to the absence of expression modeling capability. In contrast, our extended framework with expression prediction branches successfully reconstructs various facial expressions with high fidelity. Among the evaluated pretrained facial representation encoders, the PaCo-FR-based implementation exhibits superior performance, producing the most accurate and natural-looking expressive 3D faces.

\begin{table}[!tbp]
  \caption{Quantitative comparison on the NoW Validation Benchmark, with 3D face reconstruction error as the metric.}
  \vspace{-3mm}
  \label{tab:face_reconstruction_now}
  \centering
  \begin{adjustbox}{width=\columnwidth}  
  \begin{tabular}{lccccccc}
  \toprule
  \multirow{2}{*}{{Method}} & \multirow{2}{*}{Pre-train} & \multicolumn{3}{c}{{Non-Metrical\ $\downarrow$}} & \multicolumn{3}{c}{{Metrical\ $\downarrow$}} \\
  \cmidrule{3-8}
  && Median & Mean & Std & Median & Mean & Std \\
  \midrule
   MICA\cite{zielonka2022mica} & - & 0.91 & 1.13 & \textbf{0.94} & 1.06 & 1.36 & 1.18 \\
   +FaRL\cite{zheng2022general} & 20M & 0.88 & 1.19 & 1.11 & 0.95 & 1.28 & 1.18 \\
   +MCF\cite{wang2023toward} & 20M & 0.88 & 1.16 & 1.06 & 0.96 & 1.26 & 1.13 \\
  \rowcolor{green!15!gray!10} +PaCo-FR & 20M & \textbf{0.83} & \textbf{1.12} & 1.06 & \textbf{0.88} & \textbf{1.18} & \textbf{1.11} \\
  \bottomrule
  \end{tabular}
  \end{adjustbox}
  \vspace{-5mm} 
\end{table}

\subsection{Ablation Study}
\label{sec:ablation}

As illustrated in Table~\ref{tab:ablation} and Figure~\ref{fig:ablation}, our analysis reveals the impact of different components of the PaCo-FR method on the performance of downstream tasks.

\noindent\textbf{Impact of Codebook Size}. We analyzed and visualized the guidance provided by codebook space size and the priors from the belief predictor in the pre-training process, as shown in Figure~\ref{fig:ablation}. Table~\ref{tab:ablation} compares the test results on the LaPa and AFLW-19 datasets under different component configurations in the ablation experiments with a frozen backbone.
From the table, it can be observed that increasing the codebook space size from 1 to 196 benefits the model in learning more global structures and local semantic information. The visualized results indicate that with a codebook space size of 196, compared to the case with a size of 1, the model can capture more detailed facial features. When the codebook space size is 196, the pretrained model performs similarly to PaCo-FR in the face alignment task. This aligns with our expectation, as the MIM paradigm enhances the model's understanding of facial spatial structures, making its performance on face alignment tasks, which lean towards global facial information, comparable to PaCo-FR.

\noindent\textbf{Effect of Belief Predictor}. In this case, Belief Predictor needs to provide essential priors for model training, allowing it to learn more nuanced semantics within patches. Conversely, not using Belief Predictor to provide certain priors leads to confusion in the learning of model, resulting in worse performance ($3\times 196 $ tokens + random select).

In our method, there is an Incubation Stage designed for training the Belief Predictor and preventing overfitting that could collapse the learning of model. To validate this, we removed this stage, meaning Belief Predictor was continuously updated synchronously during pre-training. This approach ultimately led to a sharp decline in the model's ability to learn facial representations, resulting in poorer downstream task performance and facial generation quality.

\vspace{-3mm}
\section{Conclusion}

In this paper, we propose an end-to-end facial representation pre-training method called PaCo-FR, which is designed based on the principle of similarity of elements at the same position in aligned faces and the differences between identical elements. This method places the codebook at the image end in units of patches. Our approach is a highly efficient face pre-training method. Despite only using 2M pre-training data, the model evaluation results surpass those of FaRL~\cite{zheng2022general} with 20M data. 

{
    \small
    \bibliographystyle{ieeenat_fullname}
    \bibliography{main}
}


\end{document}